%% file: main.tex
\documentclass{article} 
\usepackage{iclr2019_conference,times}
\usepackage{graphicx}
\graphicspath{ {./images/} }

\input{math_commands.tex}

\usepackage{hyperref}
\usepackage{url}

\title{Exploring the Hyperparameter Landscape of Adversarial Robustness}

\author{Evelyn Duesterwald \\
IBM Research\\
Yorktown Heights, NY, USA\\
\And
Anupama Murthi \\
IBM Research\\
Yorktown Heights, NY, USA\\
\And
Ganesh Venkataraman \\
IBM Research\\
Yorktown Heights, NY, USA\\
\And
Mathieu Sinn\\
IBM Research Ireland \\
Mulhuddart, Dublin 15 \\
\And
Deepak Vijaykeerthy\\
IBM Research India\\
Bangalore, India \\
}

\iclrfinalcopy 
\begin{document}

\maketitle

\begin{abstract}
\input{abstract}

\end{abstract}

\input{introduction}

\input{background}

\input{autohardening}


\input{conclusions}

\bibliography{references}
\bibliographystyle{iclr2019_conference}

\input{appendix}

\end{document}

%% file: math_commands.tex
\usepackage{amsmath,amsfonts,bm}

\def\eqref#1{equation~\ref{#1}}

\def\1{\bm{1}}

\DeclareMathAlphabet{\mathsfit}{\encodingdefault}{\sfdefault}{m}{sl}
\SetMathAlphabet{\mathsfit}{bold}{\encodingdefault}{\sfdefault}{bx}{n}



%% file: abstract.tex
Adversarial training shows promise as an approach for training models that are robust towards adversarial perturbation. In this paper, 
we explore some of the practical challenges of adversarial training. 
We present a sensitivity analysis that illustrates that the effectiveness of adversarial training hinges on the settings of a few salient hyperparameters. We show that the robustness surface that emerges across these salient parameters can be surprisingly complex and that therefore no effective one-size-fits-all parameter settings exist. We then demonstrate that we can use the same salient hyperparameters as tuning knob to navigate the tension that can arise between robustness and accuracy. Based on these findings, we present a practical approach that leverages hyperparameter optimization techniques for tuning adversarial training to maximize robustness while keeping the loss in accuracy within a defined budget.

%% file: introduction.tex
\section{Introduction}

Deep learning continues to advance and reach or surpass human-level performance across tasks spread over a set of modalities such as speech~\citep{speechhumanparity} and computer vision~\citep{facehumanparity, imagenethumanparity}. At the same time, deep learning models are shown to be vulnerable to adversarial examples that are designed to misguide a trained model through small and imperceptible perturbations of the input 
data ~\citep{goodfellow2013intriguing}.

These model vulnerabilities expose a new class of security and privacy issues and spawned significant research into how to craft these perturbations and how to train models that are robust to them. Many of the earlier defense directions have resulted in only marginal increases in robustness ~\citep{ruitong2015learning,shixiang2014robustness}.  Adversarial Training~\citep{shaham2015adversarialtraining,goodfellow2013intriguing}, where adversarial examples are directly incorporated into the training process, presents one of the more promising directions that has recently shown to provide strong protections~\citep{madry2017pgm}.

In this paper, we look into some of the practical challenges of adversarial training.  The first such challenge is the need to adequately configure adversarial training protocols. In addition to common training hyperparameters, adversarial training has a few new parameters that control the significance given to adversarial perturbation during training. 
Our experiments show that the effectiveness of adversarial training hinges on adequate settings of these salient parameters. 
As is the case for many hyperparameters, adversarial training parameters are sensitive to both training data and network characteristics and one-size-fits-all settings are generally ineffective. When exploring just how sensitive adversarial training is to parameter settings, we discovered that the robustness surface that results across the parameter space can be surprisingly complex, further adding to the challenge of configuring adversarial training. 

Another practical challenge results from the often observed tension between accuracy and robustness~\citep{su2018}. The objective of adversarial training is to increase robustness, which is typically measured as the accuracy of a model's predictions on adversarial examples. However, these increases often come at the price of a decrease in accuracy on clean test examples. Recently, \cite{tsipras2019} were able to show that there, in fact, exists an inherent trade-off between accuracy and robustness. This trade-off also manifests in our experiments where 
the same parameters that are crucial to achieving robustness also tend to control the decrease in accuracy on clean test examples. 

Moreover, our experiments show that we can effectively use these salient parameters as tuning knobs to navigate the tension between robustness and accuracy. We show that by looking at adversarial training as a multi-objective tuning problem, we can leverage existing HPO (hyperparameter optimization) strategies~\citep{bergstrahpo2011,bergstrabengiohpo,snoek2012} to fine-tune parameter settings for the desired accuracy-performance trade-off.
In summary, the main contributions of this paper are as follows:

(i) A sensitivity analysis that provides new insights into the potentially complex relationship between robustness achievable through adversarial training  and its hyperparameter settings. \\
(ii) A practical approach that addresses the challenges of finding effective adversarial training hyperparameter settings through tuning. \\
(iii) An experimental evaluation of our approach that demonstrates that we can leverage HPO strategies to find hyperparameter settings that maximize robustness within a given accuracy budget.

The resulting tuning approach provides a degree of automation that enables practitioners to achieve robustness without having to reason about salient parameter settings. We believe that thereby, our approach can play an important role in making adversarial training more accessible and practical.

%% file: background.tex
\section{Background}
\label{sec:background}

Given a deep learning model $f(.; \theta)$ and a clean input $x$, whose actual label is $y$, an \emph{adversarial example}~\citep{goodfellow2013intriguing} is an input $x_{adv}$ such that: 
\[
f(x_{adv}; \theta) \neq y \land d(x, x_{adv}) \leq \epsilon
\]
Intuitively, an adversarial example $x_{adv}$ can be created by adding a small amount of noise to $x$, such that $x$ and $x_{adv}$ are close according to some distance metric $d$ but $f(x_{adv}; \theta) \neq y$. In each domain, the distance metric used varies. For instance, in the case of images various $l_p$ norms are used as they are considered a reasonable approximation of human perceptual distance. 
As commonly used to measure adversarial perturbations in images, we use $l_{\infty}$ norm, 
which is a measure of the maximum absolute change
to any pixel (worst case) 
\citep{shaham2015adversarialtraining,carlini2017l2attack}. 

Numerous techniques have been proposed for 
generating adversarial examples~(incl. ~\cite{goodfellow2014explaining,ifgsm,dezfooli2016deepfool,papernot2016jsma,carlini2017l2attack}). In this paper we focus on the Projected Gradient Descent method, a strong adversary proposed by \citet{madry2017pgm}). 


\textbf{Projected Gradient Descent (PGD)}:
PGD can be interpreted as a projected gradient descent on the negative loss function.
\citet{madry2017pgm} demonstrate on MNIST and CIFAR-10 that models adversarially trained with PGD exhibit robustness even against stronger white-box adversaries. This in-turn shows that we can build resilient defense strategies against white-box attackers by developing strong white-box attacks and then adversarially retraining the models with examples from the stronger attacks.

{\bf Adversarial Training}~\citep{goodfellow2014explaining} refers to a training protocol with the goal to harden a model against adversarial examples.
We consider a variant of Empirical Risk Minimization (ERM) proposed by~\citet{madry2017pgm}, where the aim is to minimize 
an empirical risk formulation that incorporates adversarial perturbations: 
\begin{equation*}
\min_{\theta} \rho(\theta), \mbox{~where~~} \rho(\theta) = \mathbb{E}_{(x,y) \sim D} \Bigg\{\max_{\delta \in \mathcal{S}} L(\theta; {x} + \delta, y) \Bigg\}
\end{equation*}
In the above equation $\theta$ is the set of model parameters, $\mathcal{S}$ is a set of allowed perturbations and $L(\theta; {x}, y)$ is the original loss function.
The formulation offers a unifying perspective over the existing works in adversarial robustness. Intuitively the inner maximization problem aims to find an adversarial example with high loss and the outer minimization problem tries to find the optimal model parameters such that loss due to the inner maximization problem is minimized. This is precisely the problem of training a classifier robust to the adversarial noise in the training instances.


%% file: autohardening.tex
\section{Adversarial Robustness}

We conducted a series of experiments to explore the robustness that can be achieved with adversarial training by varying hyperparameters. 
We used the adversarial trainer implementation from the open-source Adversarial Robustness Toolbox (ART) library~\citep{art}.
As proposed by ~\citet{madry2017pgm} we use PGD to generate adversarial examples during training.

We experimented with two standard image datasets: MNIST handwritten digit recognition ~\citep{mnist1998lecun} 
and SVHN street view house numbers ~\citep{netzer2011svhn}. 
We trained MNIST on the LeNet-5 network~\citep{lenet5lecun} 
and SVHN on ResNet-32~\citep{he2016resnet} since in both cases, we can achieve high levels of accuracy when training with clean training data. 

\subsection{Robustness Surface}

We first evaluated the accuracy and robustness of our two models when trained with clean training examples.
We conducted a robustness test that measures robustness as the accuracy of the model on adversarial examples generated using PGD. The degree of adversarial perturbation during example generation is controlled by the parameter $\epsilon$, which provides a bound on the maximal perturbation. In the image domain, this means that the $l_{\infty}$-norm distance between original and perturbed image has to be less than $\epsilon$. 
For MNIST we configured PGD with  $\epsilon=0.3$ ~\citep{madry2017pgm} and for SVHN we used $\epsilon = 0.03$. The results of the robustness test show that neither model is robust, yielding only 17.6\% (MNIST) and 0.4\% (SVHN) accuracy on adversarial test examples, compared to 99.8\% and 99.3\% on their clean counterparts: 

\begin{table}[h]
\centering
{\footnotesize
\begin{tabular}{| l |  l | l | l |}
\hline
\multicolumn{2}{|c|}{MNIST-LeNet} & 
\multicolumn{2}{|c|}{SVHN-ResNet} \\ 
\hline
$\emph{Acc}_\emph{test} = 99.8$ & $\emph{Acc}_\emph{adv} = 17.6$ &
$\emph{Acc}_\emph{test}=99.3$ & $\emph{Acc}_\emph{adv} = 0.4$ \\
 \hline
\end{tabular}
}
\end{table}

We next conduct a sensitivity analysis to explore how the robustness gains that can be achieved with adversarial training vary with changes in salient hyperparameter values. We focus on the hyperparameters that control the strength of the influence given to adversarial perturbation during training which include the parameters \emph{ratio} and $\epsilon$: 

\begin{table}[h]
\centering
{\footnotesize
\begin{tabular}{| l | l |}
\hline
\emph{ratio} & fraction of training examples in each batch to be replaced with their adversarial counterparts \\
\hline
$~\epsilon$ & PGD attack parameter providing an $l_{\infty}$ upper bound on the allowed perturbation \\
 \hline
\end{tabular}
}
\end{table}
To understand the influence of these parameters, we generated adversarially trained models for varying values of  \emph{ratio} and $\epsilon$ and measured both accuracy and robustness on held-out test data. We trained 50 configurations for each data set: 5 values from the $\epsilon$ dimension and the 10 \emph{ratio} values 0.1,0.2, .., 0.9,1. We repeated each of the 50 adversarial training runs 10 times and averaged accuracy and robustness measurements across these 10 runs. 
We measured the robustness of each adversarially trained model using the above PGD based robustness test with fixed $\epsilon$ values ($\epsilon=0.3$ for MNIST and $\epsilon = 0.03$ for SVHN). 
Further detail on the experiment is shown in the appendix. 
In Figure~\ref{fig:3dplots} we visualize the results in the surface plots that emerge with varying values of \emph{ratio} and $\epsilon$. We used the python \texttt{scipy.interpolate} package with cubic interpolation to generate the surface plots over a 30 x 30 grid of hyperparameters.

\begin{figure}[t]
\vspace{-0.4in}
 \centering
  \hspace{-0.6in}
    \includegraphics[width=3.3in]{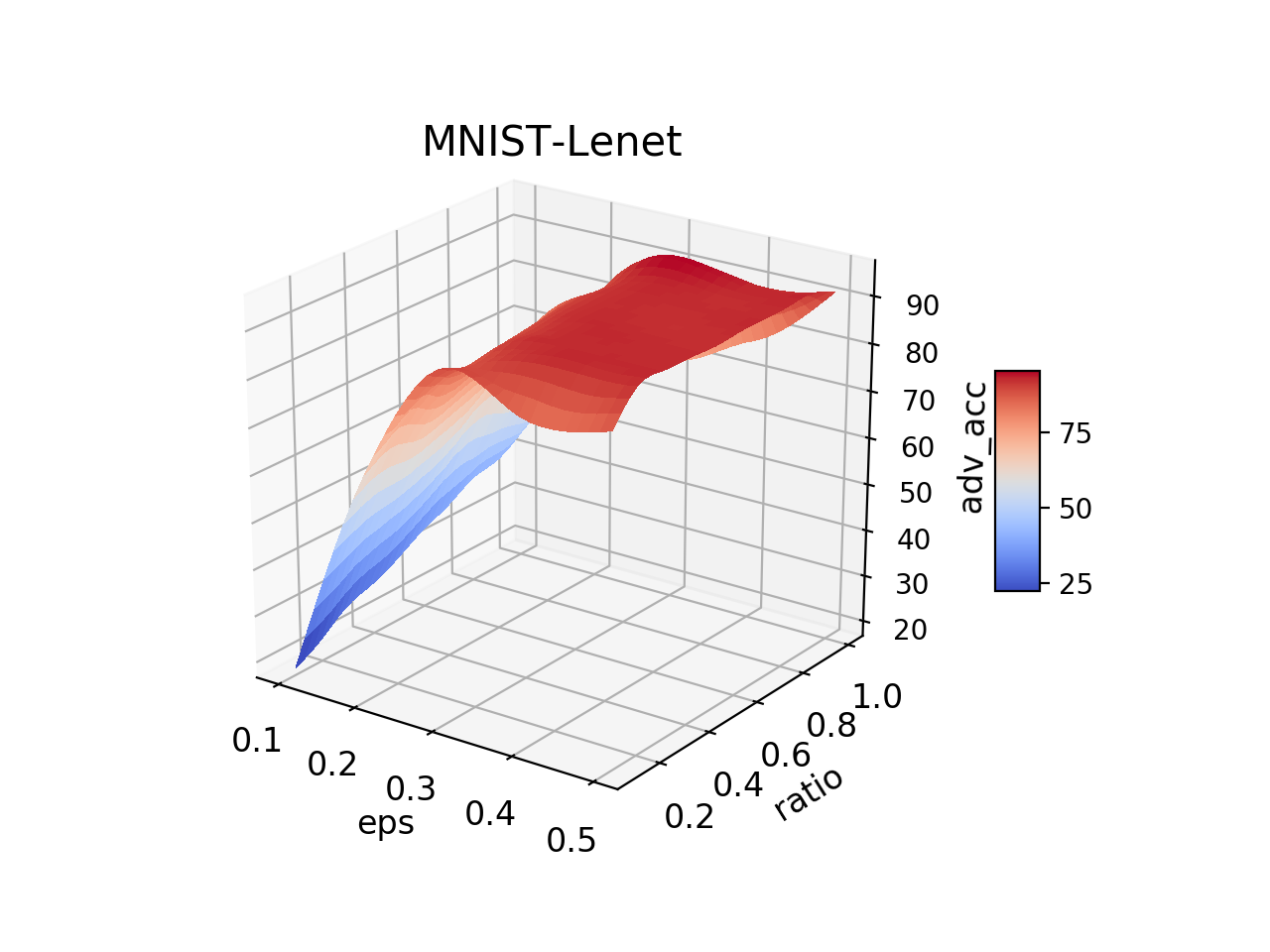} 
    \hspace{-0.5in}
    \includegraphics[width=3.2in]{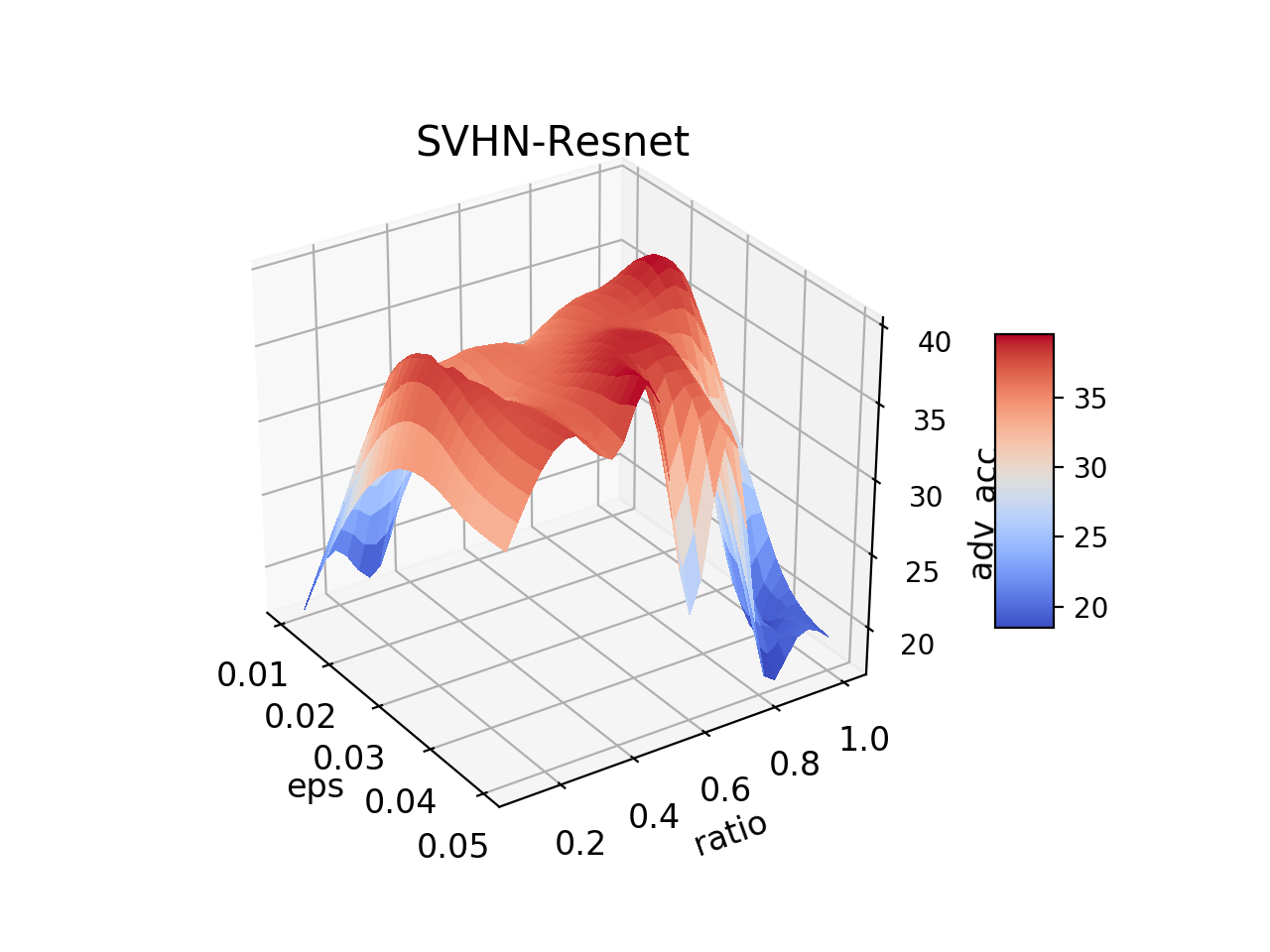} 
\vspace{-0.2in}
\caption{Sensitivity analysis illustrating the variability of robustness (adversarial accuracy) as a function of adversarial training parameters $\epsilon$ and \emph{ratio}. }
     \label{fig:3dplots}

 \centering
   \hspace{-0.8in}
    \includegraphics[width=3.0in]{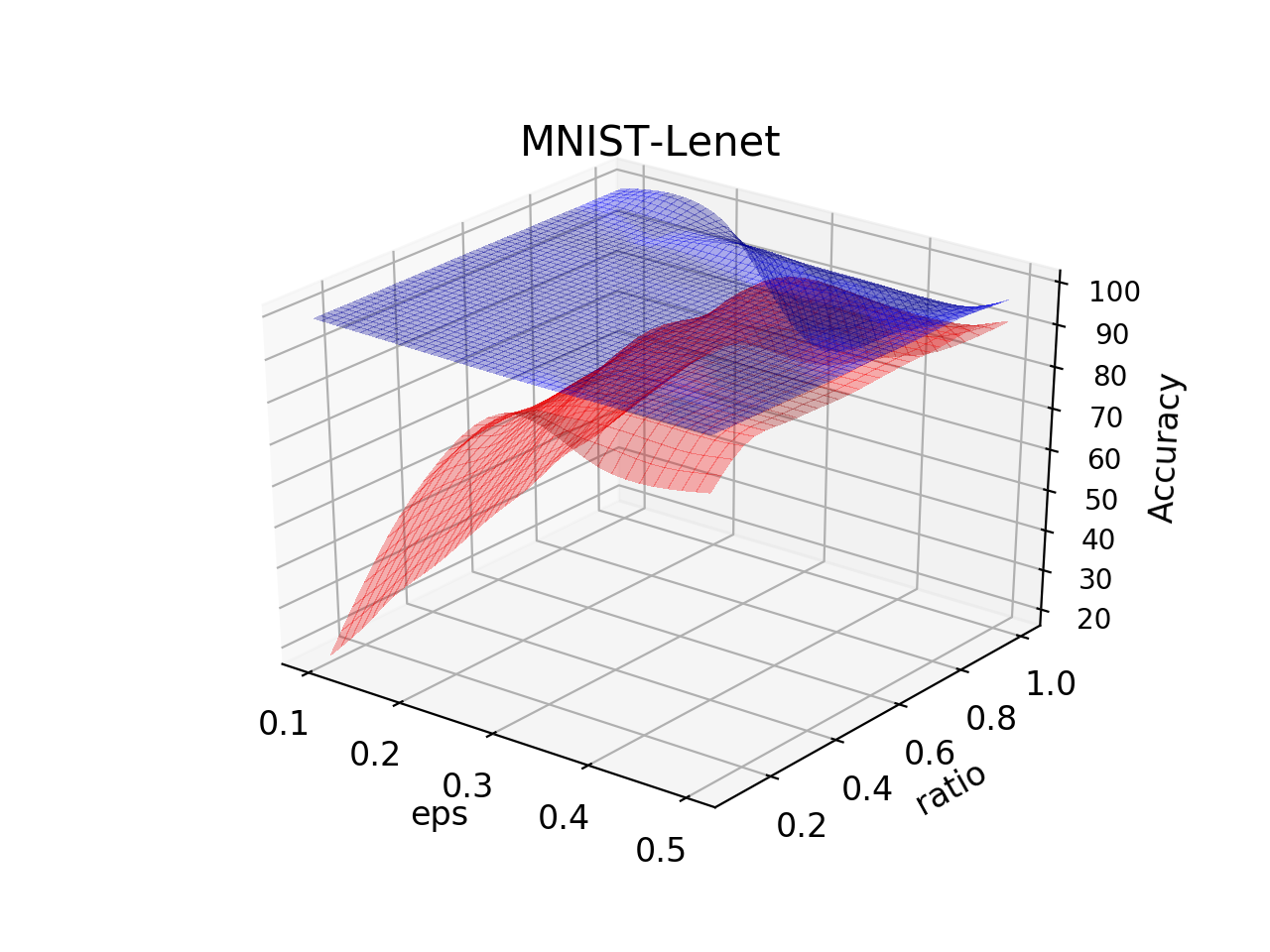} 
    \hspace{-0.2in}
    \includegraphics[width=3.0in]{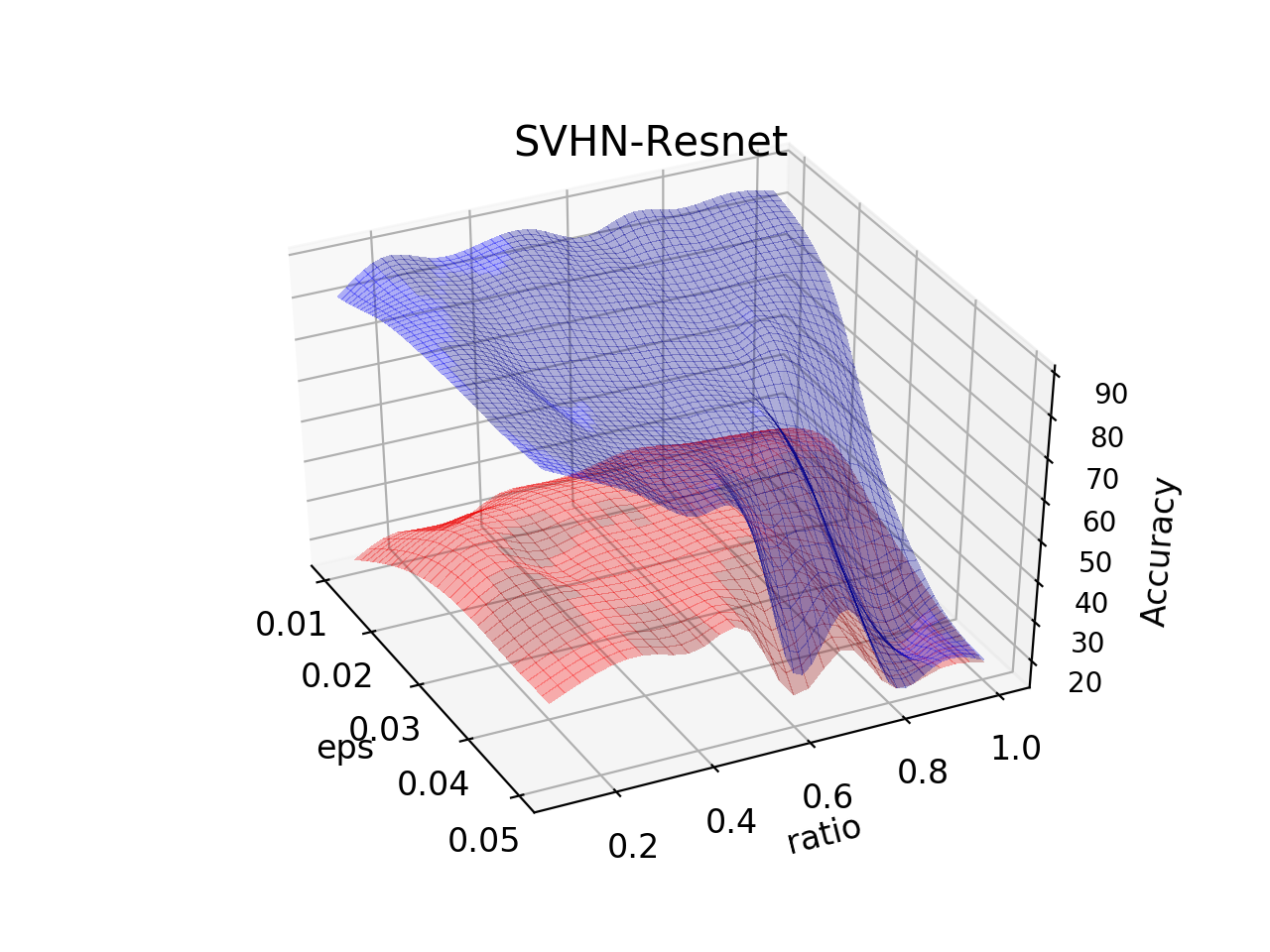} 
\vspace{-0.2in}
\caption{Accuracy surface (blue-darker color) and robustness surface (red-lighter color) that emerge across varying values of adversarial training parameters $\epsilon$ and \emph{ratio}.}
     \label{fig:3dtradeoff}
\end{figure}

Comparing the two plots shows that the shape of the robustness surface is unique to each model/data set combination as
the area with the highest robustness values is located in different areas of the parameter space in the two plots. Moreover, the surface shape itself can be surprisingly variable, especially for the more complex ResNet-32 model,  with several distinct peaks and valleys. For MNIST we observe that for $\epsilon \geq 0.3$ and ratio=1.0 the robustness of the model decreases sharply. This can be explained by the fact that we are using a model with relatively low capacity which is unable to achieve robustness under the strong adversarial loss in this parameter range (a phenomenon also observed by ~\citet{madry2017pgm}). For SVHN this effect is even more pronounced. We note that while ~\citet{buckman2018thermometer} were able to adversarially train a ResNet model on SVHN with ratio=1.0 and $\epsilon=0.047$, they used a wide ResNet architecture with width factor 4 which has the capacity to be trained against stronger adversaries. Overall, the high degree of variability both within and across plots suggests that one-size-fits-all parameter settings will be ineffective. In order to achieve maximal robustness these parameters have to be custom set.


To explore the trade-off between accuracy and robustness, we plotted the accuracy surface (accuracy on clean test examples) that emerges for the same parameter range against the robustness surface, shown in Figure~\ref{fig:3dtradeoff}. 
Both plots (MNIST and SVHN) expose the opposing trends of robustness (red) and accuracy (blue): accuracy tends to decrease with growing $\epsilon$ and \emph{ratio} whereas robustness tends to increase with growing values in these two dimensions,  up to a point, after which robustness also decreases sharply for large $\epsilon$ and \emph{ratio} values.  As observed earlier, the ResNet-32 architecture we used does not provide sufficient capacity for the SVHN data set to be fitted in the presence of a strong adversary (i.e., high $\epsilon$ and high \emph{ratio} range). However, robustness gains without significant accuracy loss can be achieved in other areas of the parameter space. Finally, we observe that the diagonal slope in the shape of the plots towards the high \emph{ratio} and high $\epsilon$ range indicates that both parameters, together, influence the shape.

\subsection{Parameter Tuning}

Figure ~\ref{fig:3dtradeoff} illustrates the challenge of finding effective adversarial training parameter settings that hit a sweet spot in the surface plots: maximizing robustness while minimizing the reduction in accuracy.
To address this tension, we are treating adversarial training as an optimization problem under control of a budget. Given a user provided budget $\beta$ on the allowable accuracy reduction relative to training on clean examples, the goal is to maximize robustness while keeping the accuracy reduction within budget $\beta$.  
Specifically, if ${\emph{Acc}}_f$ is  the accuracy that is achieved on a test set $X_{\emph{test}}$ when training on clean data, and 
${\emph{Acc}}_{f'}$ is the accuracy achieved when training adversarially, the allowable accuracy reduction has to stay within budget $\beta$:
\[  {\emph{Acc}}_{f'} (X_{\emph{test}})  >  (~ {\emph{Acc}}_f (X_{\emph{test}}) - \beta ~)
\]
To solve this problem, we leverage hyperparameter optimization (HPO) techniques~\citep{bergstrahpo2011,bergstrabengiohpo,snoek2012}. To  account for the accuracy budget $\beta$ we use $\beta$ as a filter to rule out unacceptable solutions during HPO.

We experimented with the following HPO strategies: (i) grid search, (ii) random search, and (iii) Tree-structured Parzen Estimator (TPE) ~\cite{bergstrahpo2011}. Grid search systematically explores a uniformly sampled subspace of the optimization space and returns the parameter setting for the best performing exploration. Random search randomly samples the optimization space and returns the best performing parameter setting.  TPE follows a Sequential Model-Based Optimization approach to sequentially construct models that approximate the performance of hyperparameters based on previous measurements. 

To adapt grid and random search to account for the accuracy budget $\beta$ we use $\beta$ as a filter to reject solutions with unacceptable accuracy reduction.  We adapt TPE by first running TPE unmodified while saving the results of all explored parameter settings. When TPE completes, we apply $\beta$ as a filter to eliminate solutions with unacceptable accuracy loss and return the most robust solution among the remaining parameter settings. Note that, for small $\beta$ values there may not exist an adversarially trained model that falls within that accuracy budget. 

\begin{figure*}[t]  
\vspace{-0.4in}
   \centering
    \includegraphics[width=2.2in]{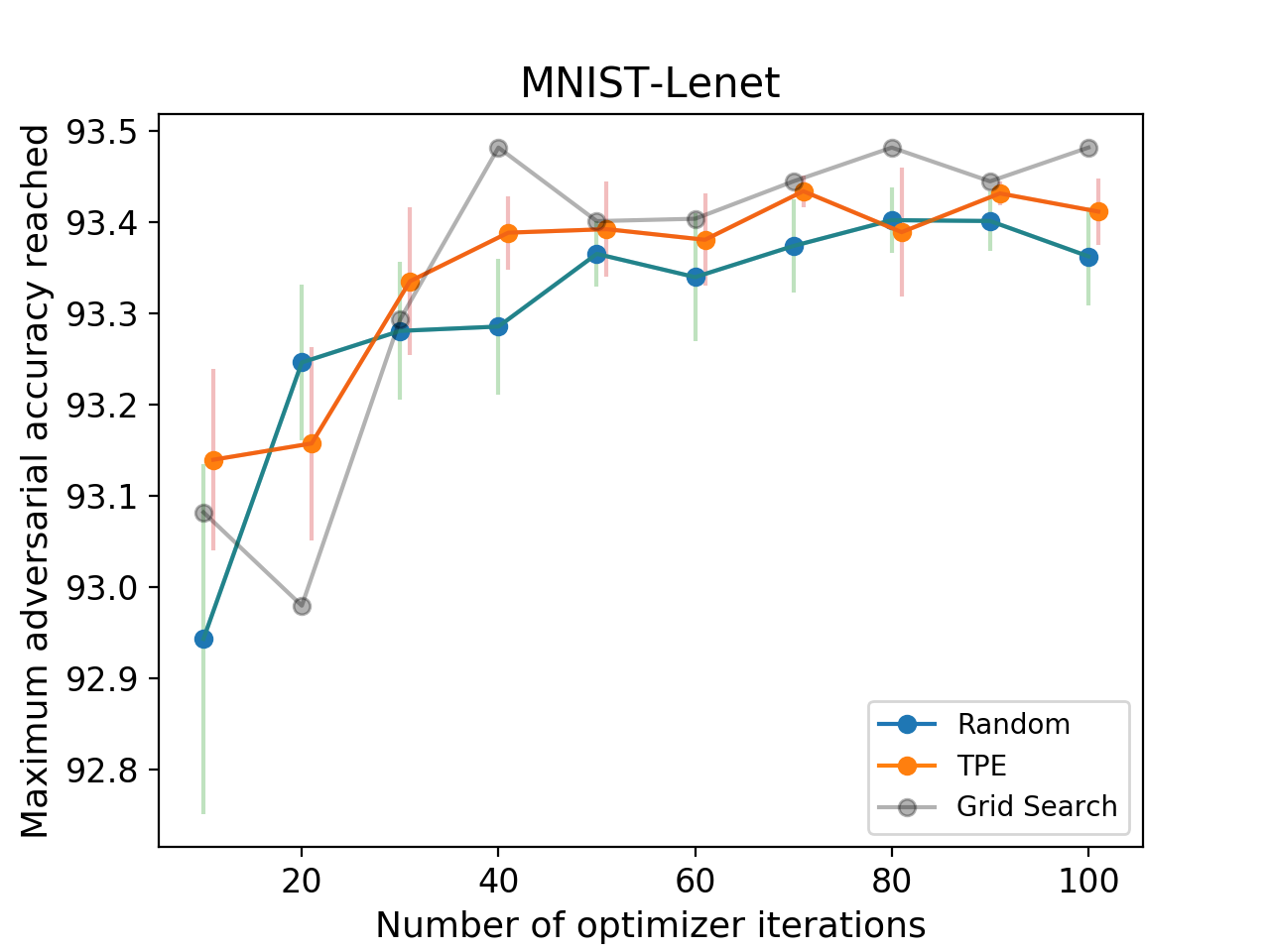}     
    \hspace{0.5in}
    \includegraphics[width=2.2in]{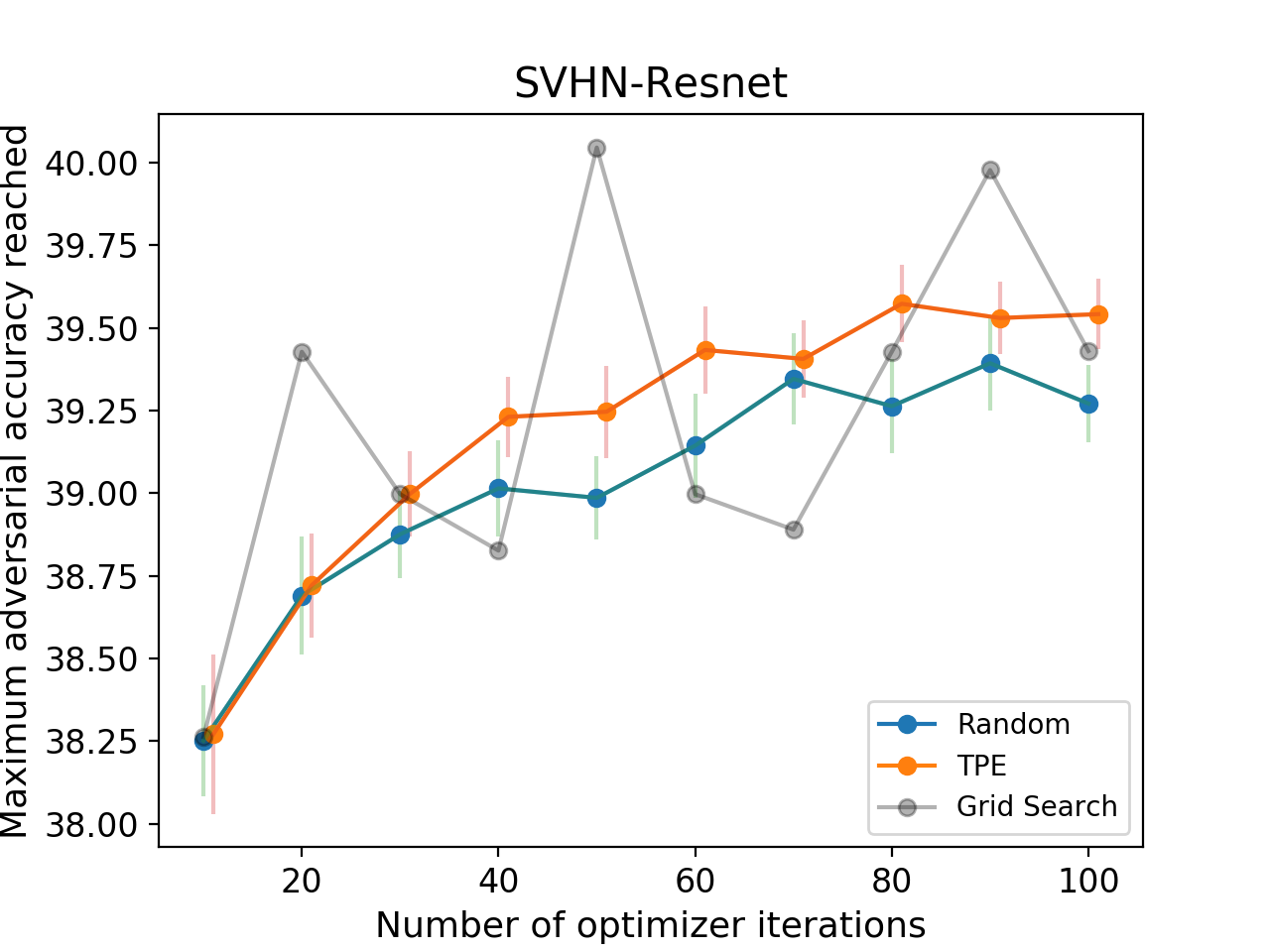} 
     \caption{Robustness achieved on the test set by grid, random search and TPE. Each random search and TPE point is averaged across 40 runs. Error bars indicate 95\% confidence intervals.}
     \label{fig:hpoplots} 
 \vspace{0.1in}
      \centering
    \includegraphics[width=2.4in]{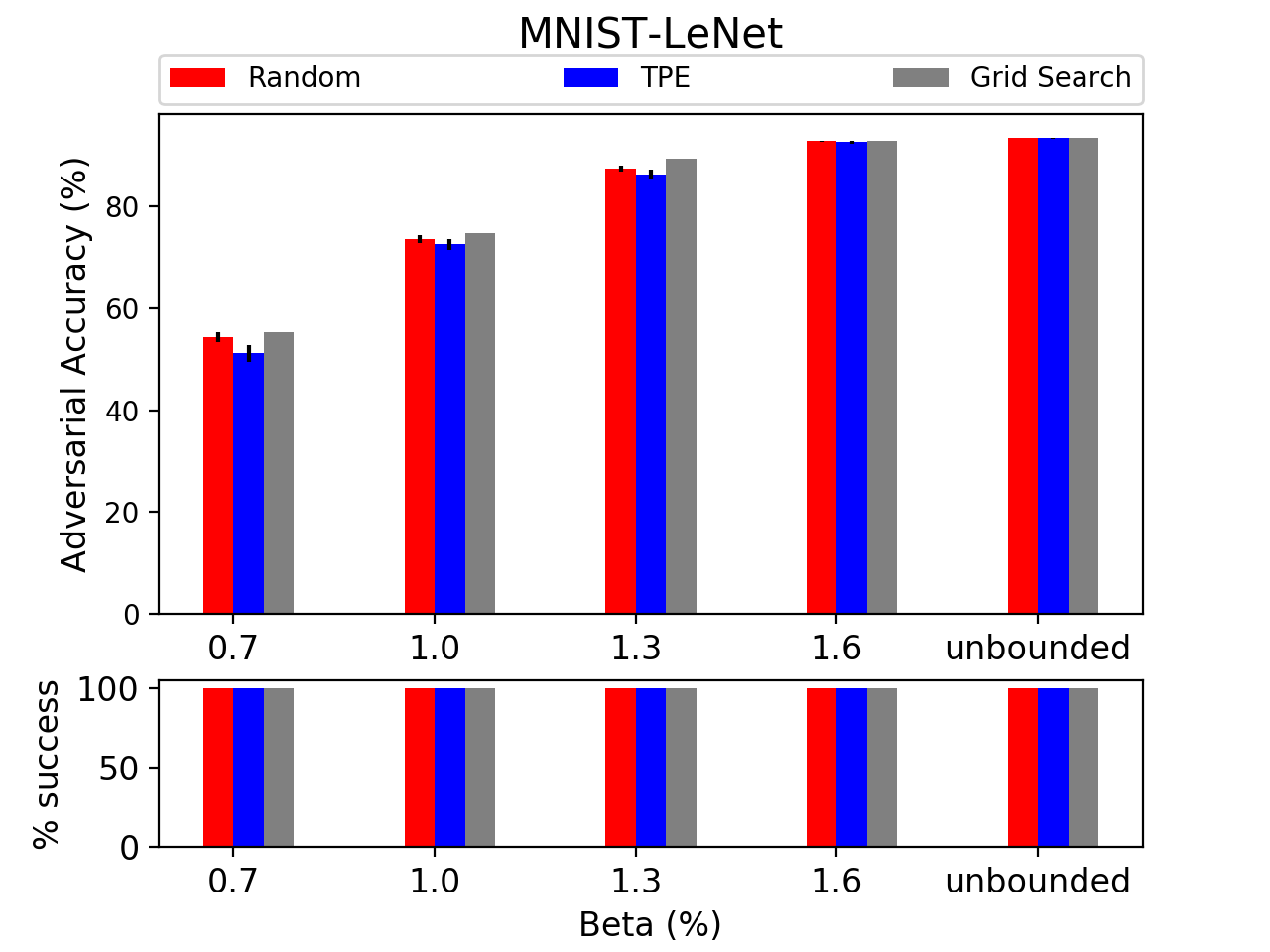} 
    \hspace{0.3in}
    \includegraphics[width=2.4in]{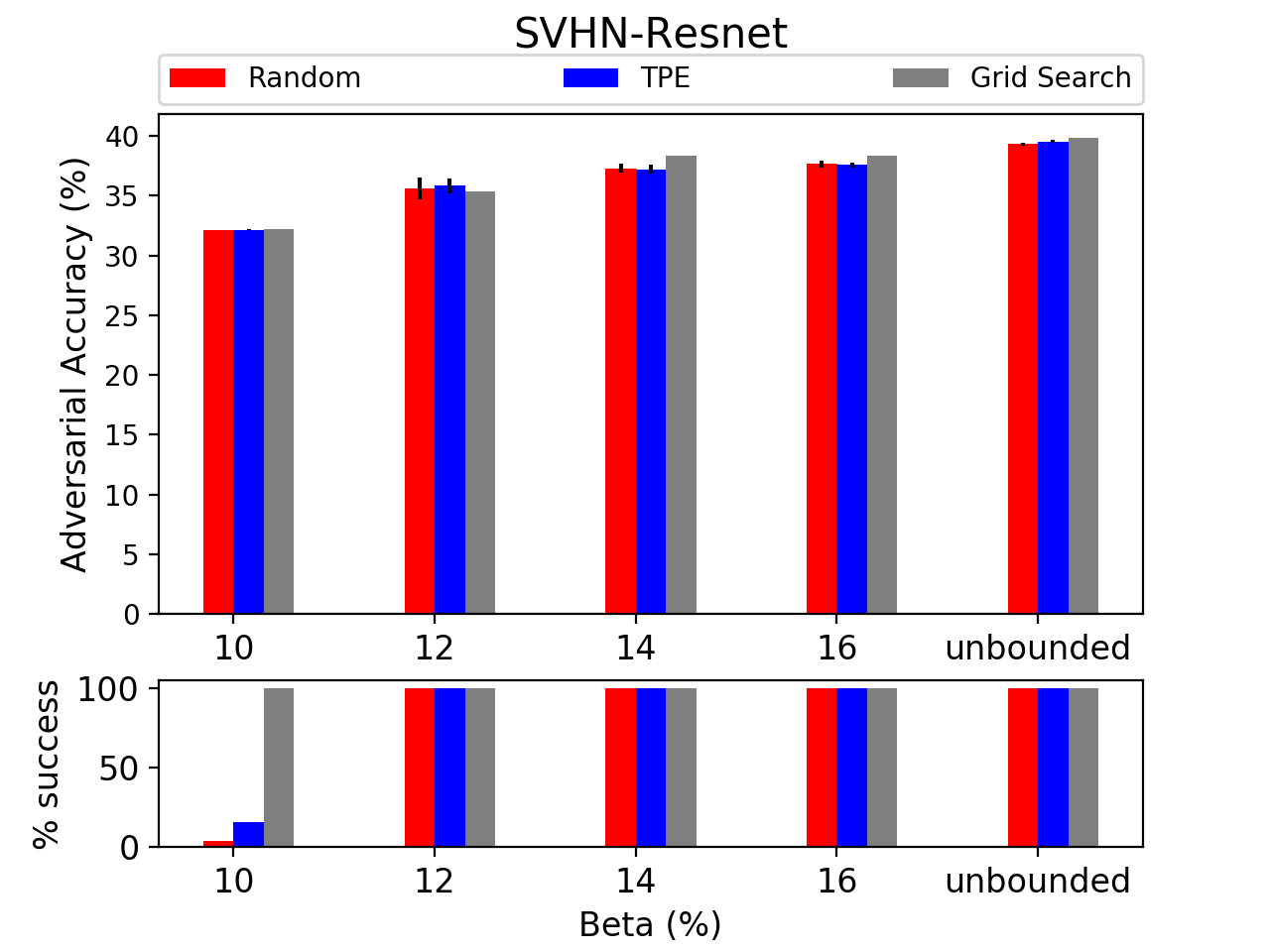}          
     \caption{Robustness achieved on the test set by various HPO techniques after 50 iterations shown for a selection of accuracy budgets $\beta$. } 
    \label{fig:hpoplots2} 
\end{figure*}

Our first set of tuning experiments evaluates the effectiveness of the three HPO techniques in finding optimal parameter settings for $\epsilon$ and \emph{ratio} without an accuracy budget constraint. We split off a 12K validation set to be used for tuning, and kept aside a 10K test set for MNIST and a 26K test set for SVHN. 
Each HPO strategy explores the parameter space at a granularity of 30 x 30  points.
We ran 100 experiments for each HPO strategy, where in experiment $i$ we ran each strategy for up to $i$ iterations (i.e., $i$ explored configurations).
Figure~\ref{fig:hpoplots} shows the achieved robustness for the three HPO techniques. 
Since both random search and TPE include random exploration we repeated each random and TPE experiment 40 times and report in Figure~\ref{fig:hpoplots} the average robustness achieved among the 40 repetitions. 

Figure~\ref{fig:hpoplots} shows that overall TPE is more effective at finding higher performing configurations in fewer iterations compared to grid and random search, except at low iteration counts, where random search performs best (especially for SVHN). Grid search performance does not grow monotonically with iterations and can be erratic due to the uniform sampling approach (the $i$ points sampled at iteration $i$ may not be included during iteration $i+1$).
Thus, while random and grid search share the same cost, random search delivers more consistent results. The cost of TPE is slightly higher since in between iterations TPE updates and applies a surrogate model. However, the costs of these inter-iteration steps are negligible compared to the cost of exploring a parameter configuration. 

All three strategies return values close to their optimum after about 50 iterations. After that point there are diminishing gains from running more iterations. 
In the following table we show the strongest results achieved at iteration $i=50$, which were found by grid search: 
\begin{table}[h]
\centering
{\footnotesize
\begin{tabular}{| l |  l l l l | l  | l l l l|}
\hline
Grid  $n = 50$ & \emph{ratio} & $\epsilon$ & $\emph{Acc}_\emph{adv}$ & $\emph{Acc}_\emph{test}$ &
Grid $n=50$ & \emph{ratio} & $\epsilon$ & $\emph{Acc}_\emph{adv}$ & $\emph{Acc}_\emph{test}$ \\
\hline
MNIST & 0.86 & 0.49 & 93.4 & 97.4 & SVHN & 0.48 & 0.049 & 40.0 & 70.9 \\
 \hline
\end{tabular}
}
\end{table}

The most robust configuration that is found for MNIST achieves strong robustness at 93.4\%. In the case of the more complex SVHN robustness surface, grid search was able to find one of the peaks  in Figure~\ref{fig:3dplots} (coordinates (\emph{ratio=0.48},$\epsilon$=0.049)), while staying clear of the area with deep robustness declines. However, this particular peak also results in a steep accuracy reduction (70.9\% compared to 99.3\% for clean training). 

Our next experiment explores how the inclusion of an accuracy budget can improve the tuning process by avoiding configurations with steep accuracy declines.  We repeated the HPO experiments for a fixed iteration count of 50 and this time incorporated an accuracy budget $\beta$.  Figure~\ref{fig:hpoplots2} shows the results of this experiment for a selection of accuracy budgets $\beta$. For each value of $\beta$, the top charts show the robustness that can be achieved under the respective accuracy budget. As in Figure~\ref{fig:hpoplots}, we repeated each random and TPE experiment 40 times and report the average adversarial accuracy achieved. 
The rightmost group of bars refers to the unbounded scenario with no accuracy budget. Since MNIST is relatively easily trained robustly, we consider very tight accuracy budgets for MNIST.


We show the results achieved for grid search below: 
\begin{table}[h]
\centering
{\footnotesize
\begin{tabular}{| l |  l l l l | l  | l l l l|}
\hline
\multicolumn{5}{| c |}{$\beta =1.6\%$}
& \multicolumn{5}{| c |}{$\beta = 14.0\%$} \\
\hline
Grid $n=50$ & \emph{ratio} & $\epsilon$ & $\emph{Acc}_\emph{adv}$ & $\emph{Acc}_\emph{test}$ &
Grid $n=50$ & \emph{ratio} & $\epsilon$ & $\emph{Acc}_\emph{adv}$ & $\emph{Acc}_\emph{test}$ \\
\hline
MNIST  & 0.49 & 0.22 & 92.7 & 98.01 & SVHN & 0.22 & 0.032 & 38.3 & 85.6 \\
 \hline
\end{tabular}
}
\end{table}

For MNIST, with $\beta=1.6\%$ grid search finds a configuration that achieves robustness close to the unbounded case (92.7\% compared to 93.4\%) with a smaller loss in accuracy. The trade-off is more interesting for SVHN where with $\beta=14\%$, grid search finds a different peak in Figure~\ref{fig:3dplots} at coordinates (\emph{ratio=0.22}, $\epsilon$=0.032) with a much lower ratio value, located in high accuracy range. The robustness is close to the robustness found with no accuracy budget (38.3\% compared to 40.0\%)  but accuracy is significantly higher at 85.6\% compared to 70.9\%. Thus, with inclusion of an accuracy budget, the HPO techniques manage to find the sweet spot in the accuracy/robustness trade-off,  even when run on a surface as complex as for SVHN.  

The bottom charts illustrate the success rate of finding configurations that fall within the accuracy budget.  For very small budgets $\beta$, random search and TPE may not always find a configuration during the 40 repetitions of the experiment. Thus, under very tight accuracy budgets, tuning with random or TPE should be repeated several times to increase the likelihood that a solution under very tight accuracy reduction bounds can be found.  


Overall, we can see that the trends in the bar charts reflect the accuracy/robustness trade-off; increases in robustness come at the price of a higher accuracy reductions. Importantly, the charts show that when adapting HPO search with an accuracy budget, we can effectively navigate this trade-off and find a satisfactory solution for a given accuracy loss tolerance.  



%% file: conclusions.tex
\section{Conclusions}
\label{conclusions}

In this paper, we studied the sensitivity of adversarial training to its salient hyperparameters.  Our experiments show that the relationship between robustness and accuracy on the one hand, and adversarial training hyperparameter, on the other, can be surprisingly complex. We devised an approach that addresses this complexity
by leveraging HPO strategies to achieve maximal robustness  within a given accuracy budget without having to reason about salient hyperparameter values.
Motivated by our results thus far we are planning to continue the exploration of the robustness surface for larger networks and tuning with additional hyperparameters. We hope that by offering a principled, automated approach we can contribute to making robust training protocols more accessible and practical.


%% file: appendix.tex

\vspace{1.0in}
\centerline{\Large {\bf Appendix}} 

Keras LeNet-5 and ResNet-32 models were trained using the IBM Watson Machine Learning service~\citep{ibmwml} with NVIDIA Tesla K80 GPUs for the smaller LeNet-5 jobs and P100 GPUs for the ResNet-32 training jobs. 
Hyperparameter settings include for MNIST-LeNet-5: batch size = 50, num epochs = 83 and for SVHN: batch size = 150, num epochs = 83. PGD was configured to run up to 40 iterations with an epsilon step size of 0.01 for MNIST and 0.007 for SVHN.